\documentclass[10pt,twocolumn,letterpaper]{article}
\pdfoutput=1
\usepackage{lipsum}
\usepackage{cvpr}
\usepackage{times}
\usepackage{epsfig}
\usepackage{graphicx}
\usepackage{amsmath}
\usepackage{amssymb}
\usepackage{macros}
\usepackage[ruled,vlined]{algorithm2e}
\usepackage{multirow}
\usepackage{afterpage}
\usepackage{subfig}
\usepackage{placeins}
\cvprfinalcopy % *** Uncomment this line for the final submission

\def\cvprPaperID{7404} % *** Enter the CVPR Paper ID here

\ifcvprfinal\fi

\begin{document}

%%%%%%%%% TITLE
\title{When to use Convolutional Neural Networks for Inverse Problems}

\author{Nathaniel Chodosh\textsuperscript{1} \hspace{0.5cm} Simon Lucey\textsuperscript{1,2}\\
\textsuperscript{1}Carnegie Mellon University \hspace{0.5cm} \textsuperscript{2}Argo AI\\
{\tt\small \{nchodosh,slucey\}@andrew.cmu.edu}}

\maketitle
%\thispagestyle{empty}

%%%%%%%%% ABSTRACT
\begin{abstract}
  Reconstruction tasks in computer vision aim fundamentally to recover an undetermined signal from a set of noisy measurements. Examples include super-resolution\cite{Dong_16}, image denoising\cite{Xie_2012}, and non-rigid structure from motion\cite{Kong_2019}, all of which have seen recent advancements through deep learning. However, earlier work made extensive use of sparse signal reconstruction frameworks (\eg convolutional sparse coding). While this work was ultimately surpassed by deep learning, it rested on a much more developed theoretical framework. Recent work by Papyan \etal~\cite{Papyan_2017} provides a bridge between the two approaches by showing how a convolutional neural network (CNN) can be viewed as an approximate solution to a convolutional sparse coding (CSC) problem. In this work we argue that for some types of inverse problems the CNN approximation breaks down leading to poor performance. We argue that for these types of problems the CSC approach should be used instead and validate this argument with empirical evidence. Specifically we identify JPEG artifact reduction and non-rigid trajectory reconstruction as challenging inverse problems for CNNs and demonstrate state of the art performance on them using a CSC method. Furthermore, we offer some practical improvements to this model and its application, and also show how insights from the CSC model can be used to make CNNs effective in tasks where their naive application fails.
\end{abstract}

%%%%%%%%% BODY TEXT
\section{Introduction}
% \begin{figure}[h]
%   \centering
%   \includegraphics[width=0.5\textwidth]{../figs/main_fig}
%   \caption{a) An example signal which is convolutionally compressible. b) An example degredation (downsampling) which preserves this convolutional structure. c) An example degradation (block diagonal) which destroys this structure.}
%   \label{fig:main}
% \end{figure}
In computer vision we often deal with signals which contain local spatial or temporal statistical dependencies. Convolution has been demonstrated to be an effective tool for modeling this type of inherent redundancy; notable examples include convolutional sparse coding (CSC) and convolutional neural networks (CNNs).  In CSC in particular, one can say that a signal is convolutionally compressible if there exists a convolutional dictionary and a sparse vector such that, 
\begin{equation}
  \label{eq:5}
  \z \approx \D\x \;\;.
\end{equation}
Here $\z$ is the signal, $\D \in \R^{N \times K}$ is the convolutional dictionary, and~$\x$ is the sparse signal such that $\norm{\x}_0 << K$. The convolutionally compressible assumption has been notably applied to many inverse reconstruction problem in vision such as image inpainting, super resolution, and trajectory reconstruction\cite{Hu_14,Yang_2008,Zhu_15}. 

Inverse problems which use the convolutionally compressible assumption can be expressed in a general form;
\begin{equation}
  \label{eq:opt}
  \arg\min_{\x} \norm{\y - \M \D \x}_2^2 + \lambda\mathbf{\Omega}(\x),
\end{equation}
where $\y$ is the degraded signal such that $\y = \M\z$, and $\mathbf{\Omega}$ provides a prior on the sparsity of $\x$. Throughout this paper we shall assume we know the degradation matrix~$\M$ a priori. 

Recently, however, CNNs have demonstrated their utility for such inverse problems\cite{Xie_2012,Dong_16}. Unlike CSC methods, however, CNNs make no explicit assumptions about the signal ($\z$) being convolutionally compressible; but are yet considered state-of-the-art for most inversion problems in vision. In this paper we want to explore if CNNs are always the correct tool for inverse problems in computer vision. In particular, we hypothesize that for inverse problems with particular types of degradation matrices ($\M$), CNNs are substantially sub-optimal in comparison to CSC inspired approaches. Notable examples include JPEG artifact removal\cite{Fu_2019}, and non-rigid trajectory reconstruction\cite{Zhu_15}. 

\noindent \textbf{Bridge between CSC and CNNs:} In this paper we shall heavily rely upon a recently established theoretical connection between CSC and CNNs by Papyan \etal~\cite{Papyan_2017}. Their work shows that a convolutional neural network (CNN) can be interpreted as approximately solving a multi-layer CSC objective. Specifically the authors state that Equation (\ref{eq:opt}) can be approximated by a feed forward CNN when $\mathbf{\Omega}$ represents a \emph{hierarchical sparsity model}. That is, instead of just assuming that $\z = \D\x$, we further assume that the sparse code $\x$ also has a convolutional structure with respect to a different dictionary, and so on. Given this observation, we can view CNNs as implicitly including the sparsity prior which is made explicit in Equation (\ref{eq:opt}). In this work we use this connection to hypothesize that for certain types of degradation matricies, the implicit convolutional assumptions of CNNs fail and in these cases CSC based methods are expected to perform better.

\noindent \textbf{When will CNNs fail?} Traditional applications of CNNs to inverse problems involve feeding the corrupted signal $\y$ into the network and predicting the original signal $\z$. As shown in the work of Papyan \etal~\cite{Papyan_2017}, this implicitly encodes the assumption that $\y$ is convolutionally compressible. This is a valid assumption for tasks such as super resolution and in painting. However, in this work, we make the observation that it is possible for $\M$ to have a structure that destroys the convolutional compressibility assumption for $\y$, despite it still holding for the original signal $\z$. We predict that for inverse problems where $\M$ has this structure, CSC based models will outperform CNNs through their ability to explicitly model the convolutional structure of the predicted signal separate from the non convolutional structure of the degradation.

\noindent \textbf{Contributions:} Our main contribution is bridging the gap between the sparse coding based analysis of CNNs and its practical application, by predicting which types of inverse problems CNNs struggle with and showing how these issues can be overcome using CSC. To validate our claim we provide four pieces of experimental evidence. First, we perform a synthetic inverse problem experiment which highlights how the structure of $\M$ affects CNN performance. Second, we show how JPEG artifact removal (AR) can be expressed as an inverse problem with non-convolutional $\M$ and achieve state of the art performance with our method. In particular we greatly outperform a recent optimization based approach~\cite{Fu_2019} to JPEG AR by explicitly modeling the non-convolutional nature of $\M$. Third, we identify non-rigid trajectory reconstruction as another inverse problem with non-convolutional $\M$, again demonstrate state of the art performance on it. With this task we also show how the CSC model can be used to design CNNs which elegantly include non-convolutional $\M$. Finally, we propose several practical contributions to the application of multi layer CSC models which improve performance in all tasks.

\section{Background and Related Work}
\label{sec:bnm}
% 1. What is sparse coding
% 2. How is it related to DL
% 3.

Traditionally, sparse coding is concerned with solving the optimization problem
\begin{equation}
  \label{eq:sparse-coding}
  \begin{gathered}
    \min_{\x} \norm{\x}_0\\
    \text{s.t.}\,\, \D\x = \y,
  \end{gathered}
\end{equation} 
where $\y$  is the signal of interest, $\D$ is known as the dictionary, and $\x$ is the sparse code. Put another way, we are searching for the most parsimonious representation of $\y$ with respect to $\D$. Unfortunately the problem as stated in (\ref{eq:sparse-coding}) is NP-Hard in general and does not allow for noise in the signal. As a result it is normally relaxed to
\begin{equation}
  \label{eq:relaxed-sparse-coding}
  \min_{\x} \frac{1}{2}\norm{\y - \D\x}_2^2 + \lambda \norm{\x}_1.
\end{equation}
which is also the Lagrangian form of LASSO regression\cite{Tibshirani_1996}. Various algorithms have been developed for this problem~\cite{Elad_2010,Daubechies_2004} and have demonstrated remarkable empirical success in denoising~\cite{Elad_Aharon_2006}, deblurring~\cite{Dong_2011}, and other image processing tasks~\cite{Mairal_2014}. One of the algorithms traditionally used to solve problem (\ref{eq:relaxed-sparse-coding}) is the Iterative Shrinking and Thresholding Algorithm (ISTA). ISTA works by repeatedly applying the formula:
\begin{equation}
  \label{eq:1}
  \x^{[k+1]} = \mathcal{S}_{\lambda} (\x^{[k]} - \D^T(\D\x^{[k]} - \y)),
\end{equation}
where $\mathcal{S}_{\lambda t}$ is the soft thresholding operator which is the proximal operator of the $\ell_1$ regularizer. $\mathcal{S}_{\lambda t}$ is given by:
\begin{equation}
  \label{eq:4}
  \mathcal{S}_{b}(x) =
  \begin{cases}
    x - b & x > b\\
    0 & b \leq x \leq -b\\
    x + b & x < -b
  \end{cases}.
\end{equation}
Recently, Papyan \etal \cite{Papyan_2017} noticed that this operator is identical to the $\ReLU$ operator except for the negative case. They then showed that if one adds a non-negativity constraint to equation (\ref{eq:relaxed-sparse-coding}) the proximal operator becomes the ReLU function and the first iteration of the corresponding thresholding algorithm becomes:
\begin{equation}
  \label{eq:2}
  \ReLU(\D^T\y - \lambda),
\end{equation}
\ie the first layer of a neural network. The authors extend this idea to a hierarchy of sparsity constraints in order to model a deep CNN. That is, they require that the sparse code $\x$ itself be sparse with respect to another convolutional dictionary, and so on. Relaxing these constraints leads to the optimization problem
\begin{equation}
  \label{eq:3}
  \min_{\{\x_i > 0\}} \sum_{i=0}^{N}\alpha_i\norm{\x_i - \D_{i+1}\x_{i+1}}_2^2 + \lambda_i\norm{\x_i}_1,
\end{equation}
where we have introduced weights for the now multiple $\ell_2$ terms and let $\y = \x_0$ for simplicity.

Applying the ideas behind ISTA to this new objective gives rise to a layered thresholding algorithm whose first pass is equivalent to a deep neural network. The authors have developed the theory of this model in several publications\cite{Sulam_2018, Sulam_2019, Aberdam_2019}, but have not applied it to the kind of large scale learning problems encountered in vision.

\noindent\textbf{Related Work:} Other authors such as Murdock \etal\cite{Murdock_18} have used the hierarchical sparsity model, but were primarily interested in its relation to component analysis and enforcing output constraints. The work of Chodosh \etal \cite{Chodosh_2018} is most similar to ours, but limited themselves to simply applying the objective to LiDAR super resolution. In contrast, this work makes a broader claim about CNNs, uses the model to predict where the CSC solution will be most effective, and demonstrates how two new tasks can be formulated as CSC problems. Furthermore we use a more theoretically sound optimization method than Chodosh \etal, and in the supplement give evidence that this leads to better performance.

Many authors have used the sparse coding model for image processing tasks. A full summary of this body of work is outside the scope of this section, but we will note some important works such as Elad \etal  which was the first to show the applicability of sparsity models for denoising\cite{Elad_Aharon_2006}, Dong \etal who introduced adaptive dictionary selection for deblurring~\cite{Dong_2011}, and Xu \etal who extended the denoising model to better deal with real world image statistics and introduced a prior-learning scheme for non-local self similar models\cite{Xu_2018,Xu_2015}.

There has also been some work on using CNNs for inverse problems which does not rely on the CSC framework. For example, Deep Image Prior by Ulyanov \etal~\cite{Ulyanov_2018} uses the implicit prior of a CNN to solve inverse problems, however they focus on a no-learning approach which does not use any data. Another notable work is that of Chang \etal~\cite{Chang_2017} which learns a projection operator for use in inverse problem optimizations. Their focus is on creating a single network which can be used in multiple inverse problems, rather than the specialized networks we are interested in here.

\section{Method}
\subsection{Applying Sparse Coding to Inverse Problems}
\label{sec:apply-sparse-coding}

Before describing our method in detail, we will first explain how sparse coding is used to solve inverse problems. Recall from the introduction that we are interested in problems of the form
\begin{equation}
  \label{eq:inverse-problem-again}
  \argmin_{\x} \norm{\y - \M\D\x} + \lambda \mathbf{\Omega}(\x).
\end{equation}
One can see that by making the substitution $\D_{1} = \M\D$ and collecting the remaining terms into $\Omega(\x)$ we can transform equation (\ref{eq:3}) into the form of (\ref{eq:inverse-problem-again}). Therefore once we have recovered the sparse codes $\{\x_i\}$ we can predict the original signal with $\hat{\z} = \D_1\x_1$. However, other authors \cite{Kong_2019} have noted that more stable recovery can be had by instead predicting
\begin{equation}
  \label{eq:4}
  \hat{\z} = \D_1\ReLU(\D_2\ReLU(\ldots\D_N\ReLU(\x_N))),
\end{equation}
which enforces the intermediate non-negativity constraints and produces better results. This is the form of prediction we use for the rest of the paper.

\subsection{Modifying the Objective}
\label{sec:solv-mlsc-object}

In Section \ref{sec:bnm} we claimed that $\ReLU(\D\y - \lambda)$ is equivalent to a neural network layer, when in fact the traditional bias vector has been replaced by the scalar $\lambda$. We will now provide a modification of the model to remedy this. Consider the following new objective function:
\begin{equation}
  \label{eq:6}
  \min_{\x_i > 0} \sum_{i=1}^{N} \frac{\alpha_i}{2}\norm{\x_{i-1} - \D_i\x_{i}}_2^2 + \norm{\bias_i \circ \x_i}_1,
\end{equation}
where we penalize each element of $\x_i$ individually by replacing $\lambda$ with the vector $\bias$, and $\circ$ represents point-wise multiplication. We can now see that the thresholding algorithm has become the standard deep learning $\ReLU(\D^\top\y - \bias)$. We note that this does \emph{not} change the expressiveness of the model, as there is a one to one correspondence between the two forms. Please refer to the supplement for full details on this and experiments showing its utility. 

\subsection{Unrolling the Optimization}
\label{sec:optim-mlsc-object}
\begin{algorithm}
  \For{$i \leftarrow 1$ \KwTo $\ell$}{
    $\x_i^{[0]} \leftarrow 0$\;
  }
  \For{$t \leftarrow 1$ \KwTo $T$}{
    \For{$i \leftarrow 1$ \KwTo $N$}{
      $\x_i^{[t]} \leftarrow \ReLU(\hat{\x}_i^{[t-1]} - \frac{1}{L_i}(\hat{\g}_{i-1}^{[t]} + \bias_i))$ 
    }
  }
  \caption{Our method for solving multi-layer CSC problems, see Equation (\ref{eq:grad}) for a definition of $\hat{\g}_i^{[t]}$}
\end{algorithm}

In their original work Papyan \etal present the LIST (layered iterative soft thresholding) algorithm for solving multi-layer CSC problems~\cite{Papyan_2017}. This algorithm is elegant, but for practical application we are interested in fast convergence, since each iteration can be quite expensive. To this end, Murdock \etal also proposed an algorithm in~\cite{Murdock_18} based on the Alternating Direction Method of Multipliers (ADMM). This was effective but relied on the assumption that $\D^\top\D = \I$ for all dictionaries, which was not satisfied in practice. This formulation and assumption was also used by Chodosh \etal~\cite{Chodosh_2018}. We avoid this issue by replacing ADMM with a recently developed convex optimization algorithm from Xu \etal~\cite{Xu_2013}.

Applying the algorithm of Xu \etal to Equation (\ref{eq:6}) gives us the iterative update:
\begin{align}
  \label{eq:update}
  \x_i^{[t]} = \argmin_{\x_i} &(\hat{\g}_i^{[t]})^\top(\x_i - \hat{\x_i}^{[t-1]}) \nonumber \\
    +& \frac{L_i}{2}\norm{\x_i - \hat{\x}_i^{[t-1]}}^2 + \norm{\bias_i \circ \x_i}_1,
\end{align}
where $\hat{\x_i}^{[t-1]} = \x_i^{[t-1]} + w_i(\x_i^{[t-1]} - \x_i^{[t-2]})$ is an extrapolation of the previous estimate of $\x_i$, $\hat{\g}_i^{[t]} = \grad f_i^{k}(\hat{\x}_i^{[k-1]})$ is the gradient of the smooth part of the objective w.r.t to the $i$th block evaluated at $\hat{\x}_i^{[k-1]}$ and $L_i$ is a step size parameter.

Equation (\ref{eq:update}) admits a closed form solution which can be derived by examining the Karush-Kuhn-Tucker (KKT) conditions for optimality. 
%We defer a full description of this to the supplement and just present the result:
This leads to
\begin{equation}
  \label{eq:9}
  \x_i^{[t]} = \ReLU(\hat{\x}^{[t-1]}_{i-1} - \frac{1}{L_i}(\hat{\g}_{i-1}^{[t]} + \bias_i)),
\end{equation}
and the gradient expression
% We will also defer computation of the gradient to the supplement as it is fairly straightforward. It works out to:
\begin{equation}
  \label{eq:grad}
  \hat{\g}_i^{[t-1]} =
  \begin{cases}
    \alpha_i\D^\top_i(\D_i\hat{\x}_i^{[t-1]} - \x_{i-1})^{[t]} \\\hspace{2mm}+\hspace{2mm}\alpha_{i+1}(\hat{\x}_i^{[t-1]} - \D_{i+1}\x_{i+1}^{[t-1]}), i < N\\
    \alpha_i\D_i^\top(\D_i\hat{\x}_i^{[t-1]} - \x_{i-1})^{[t]}, i = N
  \end{cases}.
\end{equation}
We leave the full derivation of the above results to the supplementary material.
Putting both of these equations together gives our full method for solving the multi-layer CSC objective which is summarized in Algorithm 1.
%Before we describe exactly how to solve Equation (\ref{eq:update}) for the case of the MLSC objective, we first describe an extension to the model.
% To apply this to the the MLSC objective we let $f(\x) = \sum_i{\ell} \frac{1}{2}\norm{\x_{i-1} + \D^\top\x_i}_2^2$ and $r_i(\x) = \norm{\bias_i \circ \x}_1$. The proposed algorithms are variants of block coordinate descent (BCD), that is they cyclically minimize $f$ over each block while keeping the other blocks fixed. They analyze several possible ways of updating the blocks but we will focus on just a single one as it is applied to this problem. The algorithm of interest is then summarized in Algorithm 1, for which we will now derive the explicit form of the updates.

\begin{figure*}[t!]
  \centering
  \includegraphics[width=\textwidth]{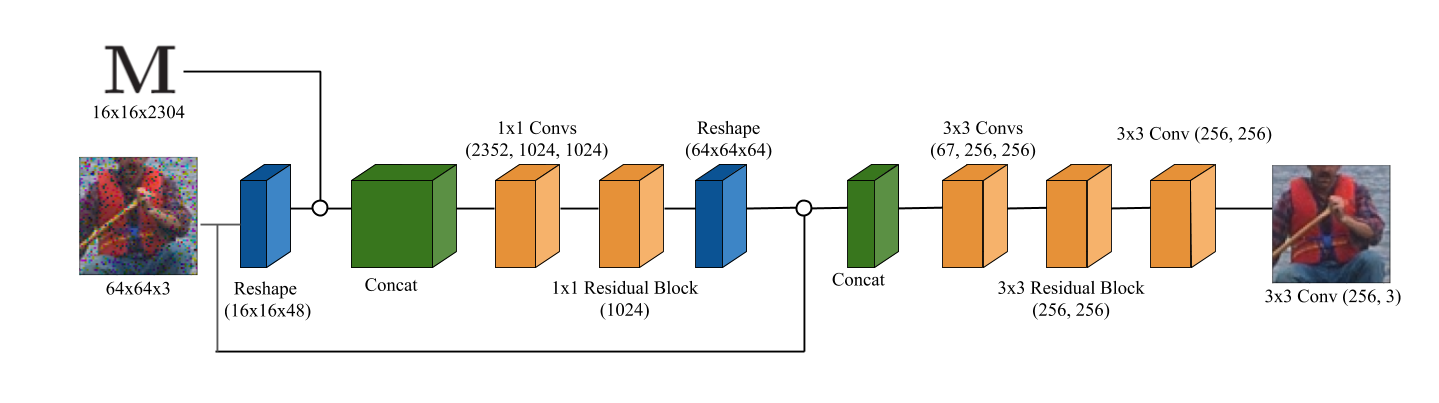}

  \caption{Our baseline CNN architecture. It is based on the JPEG Artifact Removal architecture of Zheng \etal. Due to the high dimensionality of the block-diagonal transformation we first do a patch wise encoding step. The patch-wise encoder operates on a single 4x4 patch and corresponding block of the transformation, it encodes it to a 64 channel 4x4 image which concatenated with the original image and then fed through a standard CNN. The residual blocks are of the same form as Zheng \etal}
  \label{fig:cnn-arch}
\end{figure*} 

\begin{figure}[t!]
  \centering
  \includegraphics[trim={2cm, 5cm, 1cm, 5cm},clip,width=0.5\textwidth]{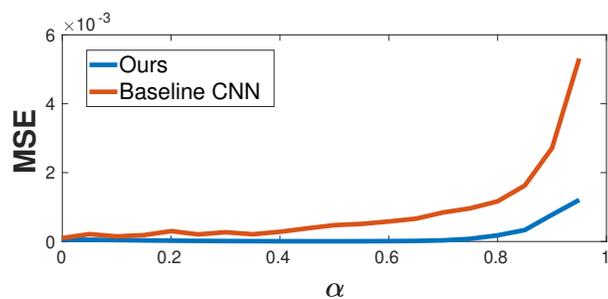}
  \caption{The results of the synthetic expperiment on BSD-500. For each value of $\alpha$ we train our CNN 5 times to convergence and then report the average of the top three validation results. In contrast we only train our model one time at $\alpha=0.5$ which explains why the performance of our model peaks there since the hyper-parameters have been learned for that value.}
  \label{fig:synthetic}
\end{figure}
\subsection{Learning Dictionaries and Parameters}
\label{sec:learn-dict-param}

In this section, we discuss how the dictionaries $\D$ are obtained in our multi-layered sparse coding pipeline. Early work in sparse coding used handcrafted dictionaries such as the discrete cosine transform or wavelet packers~\cite{Coifman_1994}, but these were ultimately outperformed by dictionaries learned by optimizing over real world datasets using algorithms such as KSVD~\cite{Aharon_2006}. Extensions of these algorithms to the multi-layer case have recently been proposed by Sulam and Elad~\cite{Sulam_2018}, which we compare against in section \ref{sec:jpeg-artif-remov}.

Let the function $\mathcal{H}^{(T)}(\y)$ represent our algorithm truncated to $T$ iterations. For simplicity we will let the output of $\mathcal{H}$ be the highest level code $\x_N$. As described in Section \ref{sec:apply-sparse-coding}, we can write our predicted signal as
\begin{equation}
  \label{eq:12}
  \hat{\z} \approx \D_1\ReLU(\ldots\D_N\ReLU(\mathcal{H}^{(T)}(\y)))).
\end{equation}
Given a reconstruction loss function $\mathcal{L}(\hat{\z}, \z)$ and $M$ training examples, we can express the goal of dictionary learning as:
\begin{equation}
  \label{eq:11}
  \min_{\{\D_i,\bias_i,L_i,w_i\}_{i=1}^N} \sum_j^M\mathcal{L}(\hat{\z}_j, \z_j).
\end{equation}
Inspired by the success of deep learning we propose to solve this objective by treating the entire inference algorithm as a black box, and optimizing the parameters with stochastic gradient descent (SGD) or one of its variants.

\section{Experiments}
\begin{table*}[t]
  \centering
  \begin{tabular}{ccccc}
    \hline\\
    Dataset & Quality & Fu\etal\cite{Fu_2019} & Zheng\etal\cite{Zheng_2018} & Ours\\\hline
    \multirow{3}{*}{BSD500} & 5 & $0.0466 \mid 26.8 \mid 0.768$ & $ 0.0381 \mid 28.5 \mid 0.853$ & $\textbf{0.0360} \mid \textbf{29.05} \mid \textbf{0.860}$\\
            & 10 & $0.0336 \mid 29.7 \mid 0.845$ & $\textbf{0.0269} \mid \textbf{32.1} \mid \textbf{0.912}$ & $\underline{0.0273} \mid \underline{31.6} \mid 0.892$\\
            & 20 & $0.0276 \mid 31.5 \mid 0.879$ & $0.0249 \mid 32.4 \mid 0.906$ & $\textbf{0.0218} \mid \textbf{33.5} \mid \textbf{0.918}$\\\hline
    Parameters ($\times 10^5$) & & 3.74 & 106.6 & .782 \\\hline
  \end{tabular}
  \caption{Results on BSD500 for reconstruction trained and tested at a specific quality factor, the numbers in each cell are RMSE, PSNR, and SSIM. Best results are in bold face, and results which are close to the best are underlined}
  \label{tab:jpeg-table}
\end{table*}
\begin{figure}
  \centering
  \includegraphics[width=0.45\textwidth]{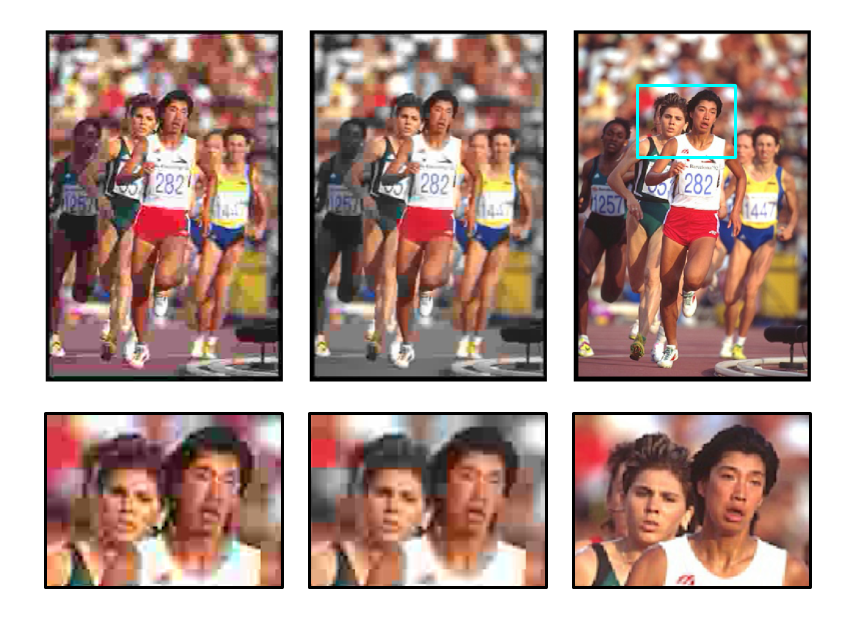}

  \caption{Top row, from left to right: an image from the Berkeley Segmentation Database compressed using JPEG, the same image degrade by our linear approximation, the original image. Bottom row, a zoom of the top images to show the blocking artifacts.}
  \label{fig:jpeg-example}
\end{figure}

\subsection{Diagonal vs Block Diagonal Inverse Problems}
\label{sec:diagonal-vs-block}

In this section we perform a synthetic experiment to show how the structure of the measurement matrix greatly affects the performance of a naive CNN approach to solving inverse problems. First we provide some intuition as to why this might be the case. The structure of a CNN encodes the belief that the input signal has a convolutional structure, that is, an individual sample is highly correlated with its neighbors. When an image is degraded by missing data or blurring, these correlations are modified, but still persist. In these cases one would expect a CNN to perform well at reconstructing the original signal and many results confirm this~\cite{Cheng_2015,Zhang_17,Urhig_2017,Ulyanov_2018}. In these tasks the corresponding measurement matrix is diagonal or convolutional.

Now consider an inverse problem where the measurement matrix is block diagonal. For simplicity we will assume that the blocks are arranged such that each 4x4 patch of the image is multiplied by an arbitrary linear transformation. We will refer to this problem as block recovery. From the perspective of solving an inverse problem, this situation is identical to inpainting or deblurring, but we would expect it to be much more difficult for a CNN.

To test this hypothesis we performed a synthetic experiment where we interpolate smoothly between inpainting and block recovery and show that the performance of a CNN degrades as we move closer to block recovery. Furthermore we will show how our proposed CSC based method is relatively unaffected by this difference in measurement matrix structure, even without retraining. Specifically we define a set of learning tasks parameterized by the scalar $\alpha$. In each task the input is a measurement matrix $\M_i(\alpha)$ and the measured image $\y_i = \M_i(\alpha)\z_i$, and the goal is to recover the original image $\z_i$. We let $\M_i(\alpha) = \M_{inpainting}(1 - \alpha) + \M_{block}\alpha$ where $\M_{inpainting}$ is a diagonal matrix which zeros out some pixels, and $\M_{block}$ is the previously described block diagonal transform. We use as data the Berkeley Segmentation Dataset (BSD-500) which provides a canonical set of high resolution images. Our baseline CNN is shown in figure \ref{fig:cnn-arch}, and to compare we use a three layer CSC model. We note that for the CNN we have to  retrain for different values of $\alpha$ but for the CSC model we simply train once at $\alpha=0.5$. The results can be found in figure (\ref{fig:synthetic}), validating our hypothesis. While interesting, this experiment wouldn't be useful if it didn't lead us to some practical application. The following experiments aim to do this by showing how two different computer vision tasks have this block diagonal structure.

\subsection{JPEG Artifact Removal}
\label{sec:jpeg-artif-remov}

In the previous section we showed how when the measurement matrix is block diagonal, we expect our CSC based method to perform better than a CNN. The synthetic experiment was contrived to demonstrate this, but we will now see how removing artifacts created by the original JPEG algorithm in fact has a similar form.

The compression algorithm operates on 8x8 blocks of the image independently. It is comprised of five steps: conversion to $\text{YC}_b\text{C}_r$ space, down-sampling of chroma channels, DCT-II transformation, rounding of DCT-II coefficients, and finally lossless compression of the remaining coefficients. The amount of rounding is controlled by a scalar parameter known as the quality factor, which varies from 1 to 100. The first three of these steps are linear transformations, and the final step is lossless and can therefore be discarded for our purposes. The fourth step creates the well known blocking artifacts, but is not linear. However as can be seen in Figure \ref{fig:jpeg-example} the majority of the artifacts are created by the frequency components which are rounded to zero. Therefore we can make an approximation to the JPEG algorithm by ignoring the rounding on frequency components which are not zeroed. This approximation is \emph{locally linear} and can be expressed as $\M_{JPEG} = \M_{zeroing}\M_{DCT}\M_{downsample}$, where we have opted to perform the experiment in $\text{YC}_b\text{C}_r$ space.

Since the algorithm processes each block separately their collective action is very similar to the previous synthetic experiment. As such a CNN would be expected to perform very poorly if applied directly to the output of the algorithm, the DCT-II coefficients. To avoid this issue, other authors choose to manually undo the the DCT and downsampling operations, effectively applying the CNN to the decompressed image. This is the approach taken by Zheng \etal~\cite{Zheng_2018} which we will use as a comparison CNN architecture. We also compare against the recent work of Fu \etal~\cite{Fu_2019}, who also chose to apply convolutional sparse coding to this problem. However, there are several key differences between our approaches: First Fu \etal bases their architecture on \emph{single layer} CSC. Second, Fu \etal applies LISTA to the objective, resulting in an network which is inspired by an optimization instead of explicitly performing it. Finally and most importantly, Fu \etal do not model the JPEG degradation and instead use the decompressed image as their input. As a result of these differences we find that our method uses far fewer parameters and achieves significantly better results. These findings are consistent with our hypothesis that explicit modeling of the measurement matrix is key when dealing with non-convolutional corruption.

We perform all experiments on the BSD-500 dataset for a range of quality factors. We also perform all experiments on the full $\text{YC}_b\text{C}_r$ image instead of on only the luminance channel. Since the method of Fu \etal~\cite{Fu_2019} was designed for only a single channel input, we double the number of features in each convolution to compensate for the larger input.

\begin{figure}
  \centering
  \includegraphics[width=0.2\textwidth]{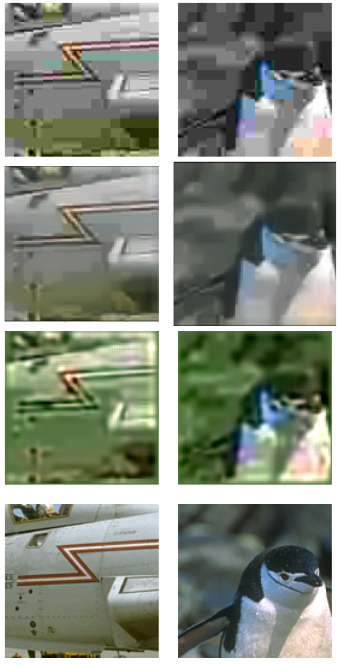}
  \vspace{1mm}
  \begin{tabular}{c|c|c}
    \hline
    Quality Factor & Ours & Sulam \etal\\\hline
    5 & \textbf{0.039} & 0.080\\\hline
    25 & \textbf{0.069} & 0.11\\\hline
  \end{tabular}\\
  \caption{From top to bottom: image degraded at QF 5, our method's output, the method of Sulam \etal, the target image. The poor performance of the method of Sulam \etal can be explained by the fact that these methods are very sensitive to the choice of parameters. The dictionary is learned with one set of parameters but there is no clear way to modify them once we introduce the JPEG degradation. The table reports RMSE for our method and Sulam \etal at quality factors 5 and 25}
  \label{fig:sulam-comp}
\end{figure}

In the first experiment we fix a quality factor and train each method to predict the original image from the compressed one. The results are show in table \ref{tab:jpeg-table}, which demonstrates that when the measurement matrix is closer to block diagonal our method achieves state of the art performance at quality factors 5 and 20 all while using much fewer parameters than the next closest model. Furthermore we are able to outperform both the deep sparse coding method and traditional multi-layer CSC by a wide margin at all quality factors. This is shown in the table as well as in the results of figure \ref{fig:sulam-comp}. That figure shows the results of a similar experiment except that we use identitcal model parameters to Sulam \etal, to demonstrate the effectiveness of learning the hyper-parameters directly from the data. Finally we performed an experiment where we mismatch the training and testing quality factors in order to measure the generalization of each model. Figure \ref{fig:jpeg-graph} shows the results of this experiment which demonstrate that due to our method specifically modeling the degradation, it outperforms all others by a wide margin when there is a mismatch between testing and training degradation.

\begin{figure}
  \centering
  \includegraphics[width=0.45\textwidth]{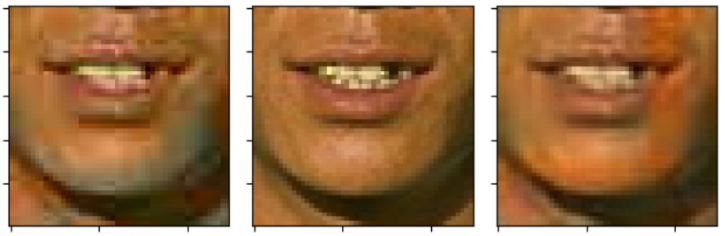}
  \includegraphics[width=0.45\textwidth]{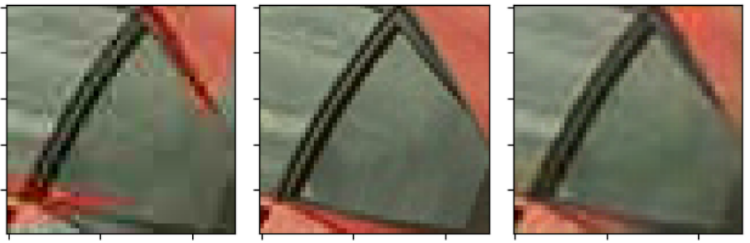}
  \caption{Some example artifact removal results from BSD-500, zoomed in to highlight the detail. From left to right: Degraded input image, target image, our result}
  \label{fig:jpeg-graph}
\end{figure}

\begin{figure}
  \centering
  \includegraphics[trim={1cm, 5cm, 1cm, 5cm},clip,width=0.5\textwidth]{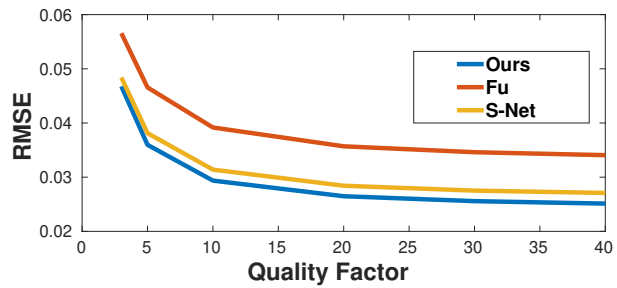}
  \caption{A comparison of our method and that of Zheng \etal on the JPEG artifact removal task. Note that as the degradation becomes more sever, that is the blocks of the measurement matrix become less diagonal, we see the same kind of fall off in performance as we did in the synthetic experiment.}
  \label{fig:jpeg-graph}
\end{figure}

\subsection{Non-Rigid Trajectory Reconstruction}
\label{subsec:non-rigid-trajectory}
In non-rigid trajectory reconstruction, one seeks to recover the 3D trajectories of points from their 2D projections. Zhu \etal first proposed convolutional sparse coding for this problem in \cite{Zhu_15}. They demonstrate large performance gains through learning the filters directly from motion-capture data. However this requires manual tuning of the sparsity parameter which can be difficult to get right. In contrast, our method learns this parameter without tuning. Before presenting these results we will first give a brief description of the problem.

We refer the reader to Zhu \etal~\cite{Zhu_15} for a description of how the trajetory reconstruction problem can be phrased as a sparse coding one, and present here just the equations for the input signal and measurement matrix.
\begin{gather}
  \label{eq:traj-sc}
  \M =
  \begin{bmatrix}
    \M_1 & &\\
    & \ddots &\\
    && \M_F
  \end{bmatrix}, \y =
  \begin{bmatrix}
    \y_1\\\vdots\\\y_F
  \end{bmatrix}\\
\end{gather}
where the $\M_i, \y_i$ encode the weak perspective cameras and 2D points. They demonstrate that learning the dictionary $\D$ on real motion capture data yields very impressive results which are still the state of the art on this task. Using the 3D ground truth points as supervision, we can train our end to end multi-layer CSC method for this task using the same formulation as described in section \ref{sec:learn-dict-param}, except now our convolutions are in time instead of space.

\begin{table}[h]
  \centering
  \begin{tabular}{c|c|c}
    & NRMSE & RMSE\\\hline
    Zhu \etal & 0.025& 35.91\\\hline
    14 Layer Trajectory CNN & 0.0254& 39.98\\\hline
    Ours 3 Layer & \textbf{0.0179}& \textbf{27.16}\\\hline
  \end{tabular}\\
  \caption{Results of multi-layer CSC (ML-CSC) on the trajectory reconstruction problem. We report Normalized RMSE in addition to RMSE, to better compare sequences of different scales. Please see \cite{Zhu_15} for a description of this metric.}
  \label{tab:traj-res}
\end{table}
\begin{figure}[htt!]
  \centering
  \includegraphics[width=0.48\textwidth]{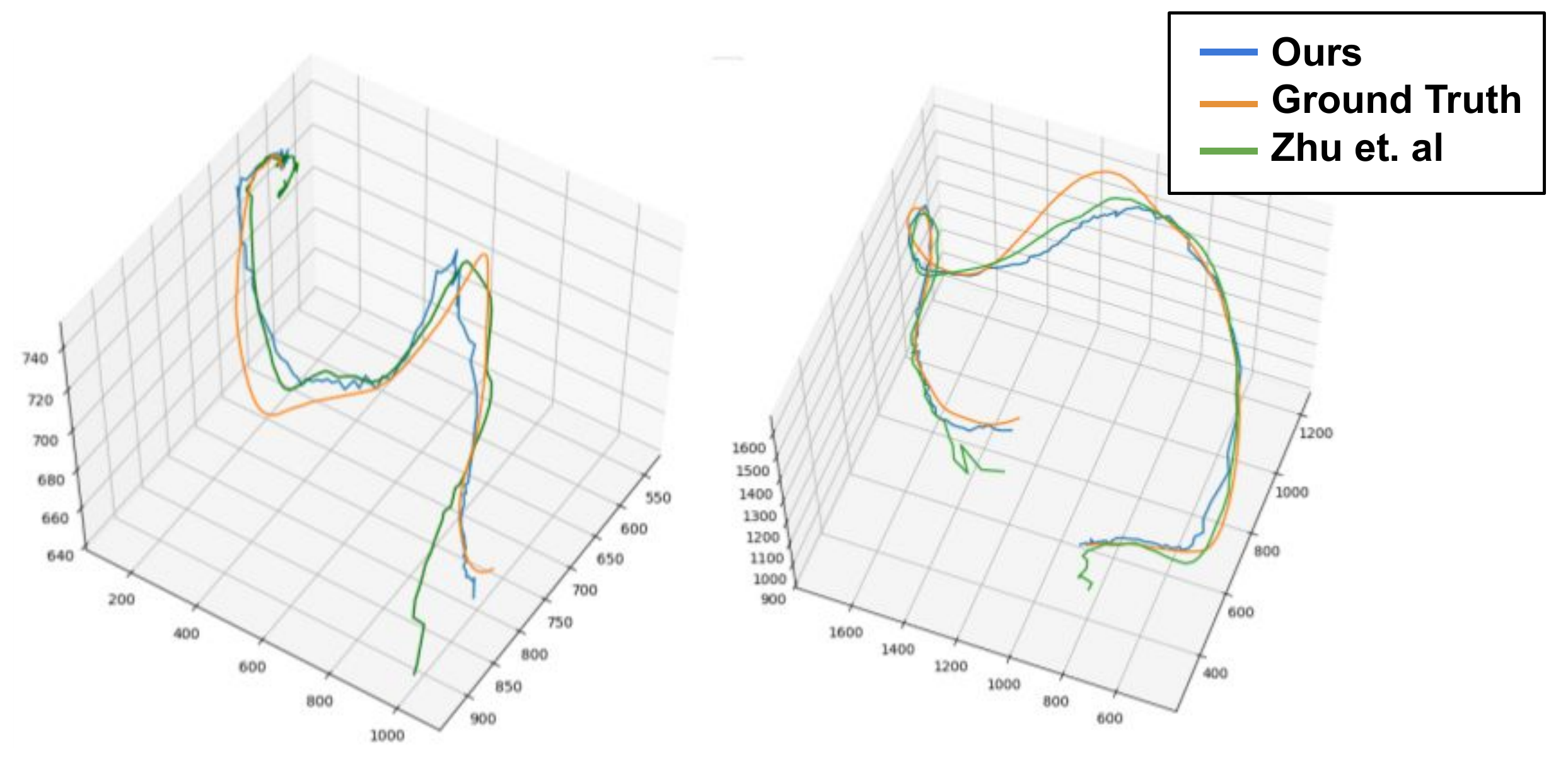}
  \caption{Some example trajectories and our reconstruction results}
\end{figure}
\begin{figure}[hbt!]
  \centering
  \includegraphics[width=0.4\textwidth]{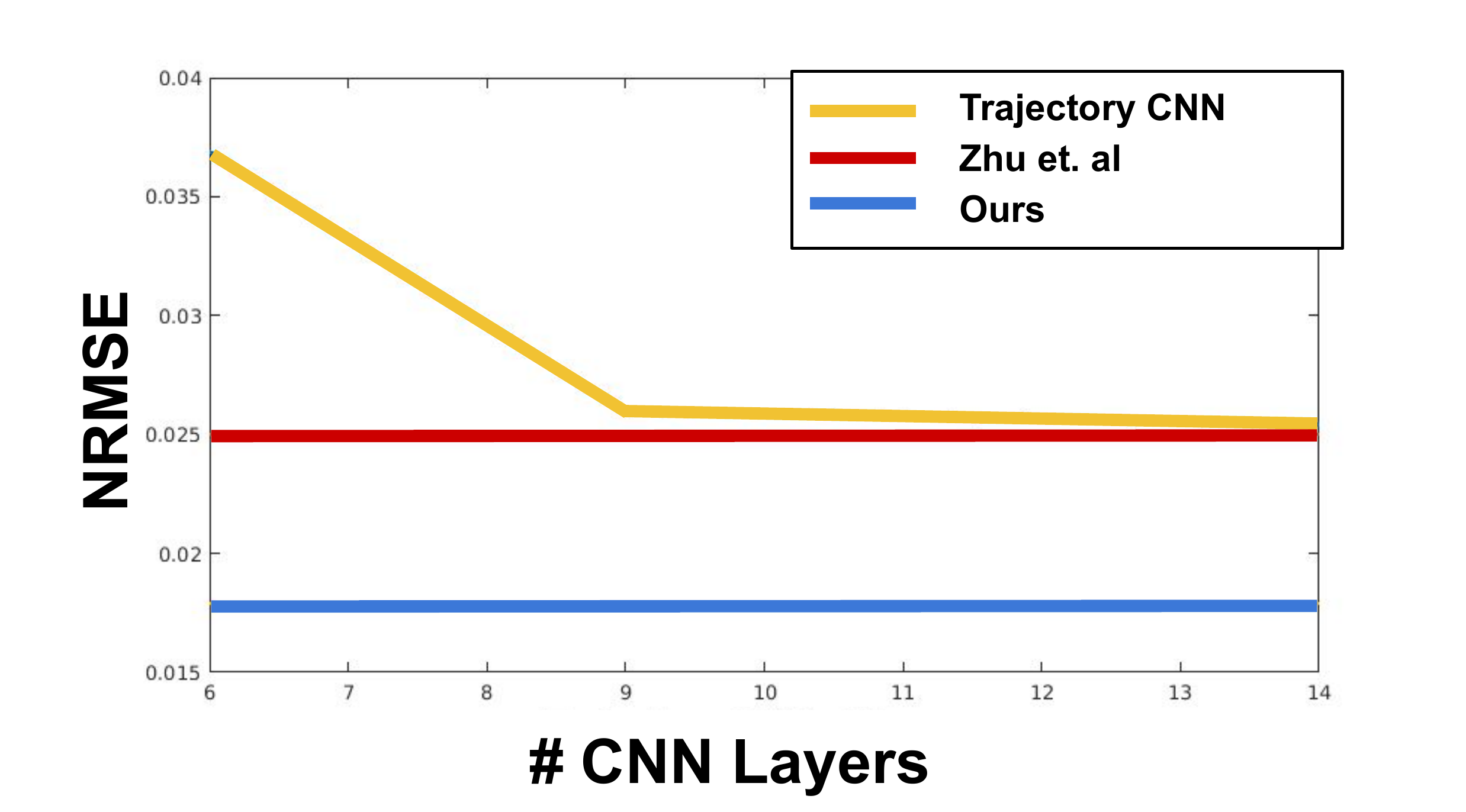}
  \caption{Performance vs Number of layers for our proposed Trajectory CNN}
  \label{fig:traj-cnn}
\end{figure}
\begin{figure*}[htb!]
  \centering
  \subfloat[text][Fu \etal: 0.0211$\mid$33.494$\mid$0.911]{
    \includegraphics[width=0.23\textwidth]{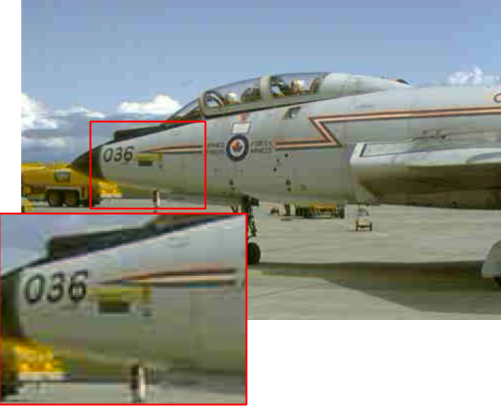}}
  \subfloat[text][S-Net: 0.0190$\mid$34.4$\mid$0.930]{
    \includegraphics[width=0.23\textwidth]{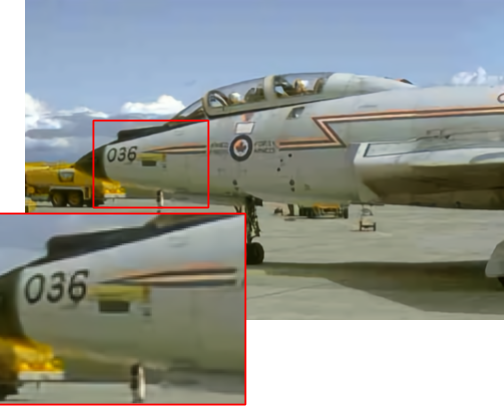}}
  \subfloat[text][Ours: 0.0150$\mid$36.5$\mid$0.946]{
    \includegraphics[width=0.23\textwidth]{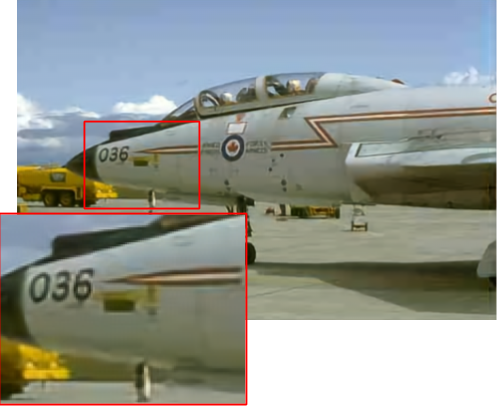}}\\
  \subfloat[text][Ground Truth]{
    \includegraphics[width=0.23\textwidth]{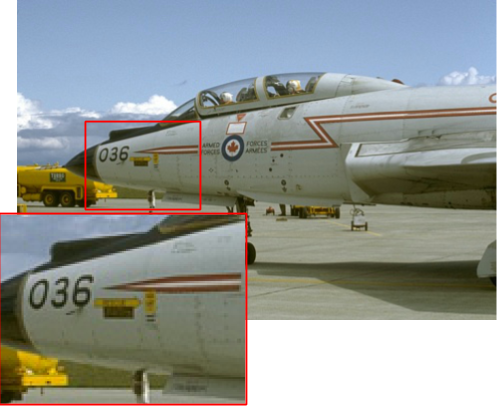}}
  \subfloat[text][JPEG]{
    \includegraphics[width=0.23\textwidth]{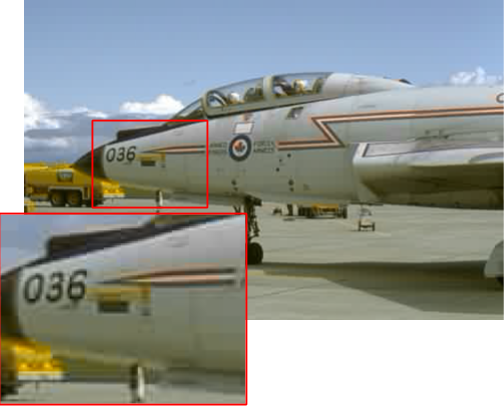}}    
  \caption{Qualitative results on BSD500. The captions contain RMSE, PSNR, and SSIM measures for each image}
  \label{fig:jpeg_example}
\end{figure*}

In our experiments we replicated the setup of Zhu \etal on the CMU motion capture dataset (mocap.cs.cmu.edu). We took each sequence, resampled them all to 30fps, created a synthetic orthographic camera which orbits the center of mass of the tracked points at a rate of $\pi/s$, and then cut the tracks into 150 frame sequences to form individual training examples. For validation, we hold out the sequences of subjects 14, 94, and 114, and use the rest for training. We can see from the results in Figure \ref{tab:traj-res} that we are able to outperform those results by a wide margin.

We do not show results for a naive CNN on this task since they fail to produce reasonable predictions. This is in line with our hypothesis that inverse problems such as these are not amenable to simple CNN approaches. Instead we will show how we can derive an modification of the input from the multi layer CSC model that allows a simple CNN to achieve reasonable results on this task.

Recall from section \ref{sec:optim-mlsc-object} that the first pass of our multi-layer CSC algorithm is equivalent to a feed forward CNN. Specifically we have that on the first iteration:
\begin{equation}
  \label{eq:1}
  \x_i^{[0]} = \ReLU(\frac{1}{L_i}\D_i^\top\x^{[0]}_{i-1} + \bias_i).
\end{equation}
However, we also have from section \ref{sec:apply-sparse-coding} that when we have a measurement matrix $\M$ then the first dictionary $\D_1$ should be replaced by $\M\D_i$. This means that the first layer's first iteration is given by:
\begin{equation}
  \label{eq:2}
  \x_1^{[0]} = \ReLU(\frac{1}{L_i}\D_i^\top\M^\top\y + \bias_1).
\end{equation}
This leads to the observation that the input to the network is effectively $\M^\top\y$ instead of simply $\y$. We note that this is similar to how in the JPEG AR task, the input to the CNN was not the DCT coefficients but instead the decompressed image. Since the DCT transform is orthogonal, performing the inverse compression is very close to multiplying $\y$ by $\M^\top$. In general the product $\M^\top\y$ does not contain all of the information required to reconstruct $\z$, as we saw in the JPEG task as well. But, we will now see that in the non-rigid trajectory reconstruction task enough of the information is preserved that it allows a CNN to perform reasonably.

To give some intuition of why we expect this to work, let $\p_i$ be the original 3D point which is projected to $\y_i$ by orthogonal camera $\M_i$. Since the trajectories are smooth in time, the $\p_i$ have a convolutional structure appropriate for a CNN. This structure is destroyed when multiplied by $\M_i$. However, we note that $\M_i^T\M_i$ has the effect of back projecting the points to the 3D space, but with the depth set to 1. Since the cameras are smooth in time as well, the resulting trajectory is a distorted version of the original which is smooth. Thus we expect a CNN to perform well on this modified input. Our prediction is confirmed by the results in table \ref{tab:traj-res}. This network is able to achieve similar results to the CSC approach but requires many more layers to make up for the lack of iterations as shown in Figure \ref{fig:traj-cnn}.

\section{Conclusion}
In this work we have used multi-layer CSC and its relationship with CNNs to predict when CNNs will perform poorly on linear inverse problems. We tested this hypothesis on a synthetic task and two real world ones and in all cases we found that when faced with measurement matricies which do not posses a convolutional structure, a naive CNN approach fails, and modification of the input is required for good performance. Specifically we achieve state of the art performance in both JPEG artifact reduction and non-rigid trajectory reconstruction. Furthermore, we used this observation to develop a CNN which can effectively solve the non-rigid trajectory reconstruction task, which previously had not seen a deep learning based approach. Additionally we made practical contributions in the form of 1) modifying the MLSC objective into a weighted version 2) finding an optimization method which converges quickly and does not require any assumptions about the dictionary.

{\bf Acknowledgments:} This work was supported by the CMU Argo AI Center for Autonomous Vehicle Research.
\FloatBarrier
{\small
\bibliographystyle{ieee_fullname}
\bibliography{ms}
}
\end{document}